\documentclass[10pt,twocolumn,letterpaper]{article}

\usepackage{iccv}
\usepackage{times}
\usepackage{epsfig}
\usepackage{graphicx}
\usepackage{amsmath}
\usepackage{amssymb}
\usepackage{booktabs}
\usepackage{multirow, makecell}
\usepackage{subfloat}
\usepackage{subcaption}

\usepackage[breaklinks=true,bookmarks=false]{hyperref}
\usepackage[capitalize]{cleveref}
\iccvfinalcopy 

\def\iccvPaperID{****} 

\ificcvfinal\fi

\begin{document}

\title{Lformer: Text-to-Image Generation with L-shape Block Parallel Decoding}

\author{
\centerline{Jiacheng Li$^1$,\;Longhui Wei$^{2,3}$,\;ZongYuan Zhan$^2$,\;Xin He$^2$,\;Siliang Tang$^1$,\;Qi Tian$^2$,\;Yueting Zhuang$^1$}\\
\centerline{$^1$Zhejiang University\quad $^2$ Huawei Cloud\quad $^3$University of Science and Technology of China}\\
\centerline{\tt\small{\{jiachengli, siliang, yzhuang\}@zju.edu.cn, weilh2568@gmail.com}}\\
\centerline{{\tt\small {\{zhanzongyuan, hexin80, tian.qi1\}}@huawei.com}}\\
 }

\maketitle
\ificcvfinal\thispagestyle{empty}\fi

\begin{abstract}
Generative transformers have shown their superiority in synthesizing high-fidelity and high-resolution images, such as good diversity and training stability. However, they suffer from the problem of slow generation since they need to generate a long token sequence autoregressively. To better accelerate the generative transformers while keeping good generation quality, we propose Lformer, a semi-autoregressive text-to-image generation model. Lformer firstly encodes an image into $h{\times}h$ discrete tokens, then divides these tokens into $h$ mirrored L-shape blocks from the top left to the bottom right and decodes the tokens in a block parallelly in each step. Lformer predicts the area adjacent to the previous context like autoregressive models thus it is more stable while accelerating. By leveraging the 2D structure of image tokens, Lformer achieves faster speed than the existing transformer-based methods while keeping good generation quality. Moreover, the pretrained Lformer can edit images without the requirement for finetuning. We can roll back to the early steps for regeneration or edit the image with a bounding box and a text prompt.
\end{abstract}

\section{Introduction}
Recent years have seen significant progress in the task of text-to-image~(T2I) generation. The fidelity, diversity, and text relevance of the synthesized images are greatly improved with large-scale models~\cite{Transformer, unet} and datasets\cite{LAION400, CC}. Generative transformers
\footnote{The early works of this approach are autoregressive models, whose name is better known by the community. But there are non-autoregressive models and discrete diffusion models later, so we call them `generative transformers' appropriately.}
~\cite{DALLE, Cogview1, Make_A_Scene, NUWA, RQVAE, Parti, Imagebart, cogview2, VQDIFF} are one of the popular generative methods receiving growing interest. Generally, they consist of two stages:  Firstly, it uses a Vector Quantized Variational AutoEncoder~(VQ-VAE)~\cite{VQVAE} to compress an image into discrete image tokens, which can be decoded back into an image at inference time; Secondly, it learns to generate image tokens from the text input with a transformer model. The transformer architecture is flexible for scaling up, and the maximum likelihood estimation~(MLE) ensures better training stability and diversity compared with previous GAN-based models~\cite{AttnGAN, DM-GAN, DF-GAN}.

\begin{figure}[tbp]
    \centering
    \includegraphics[width=\columnwidth]{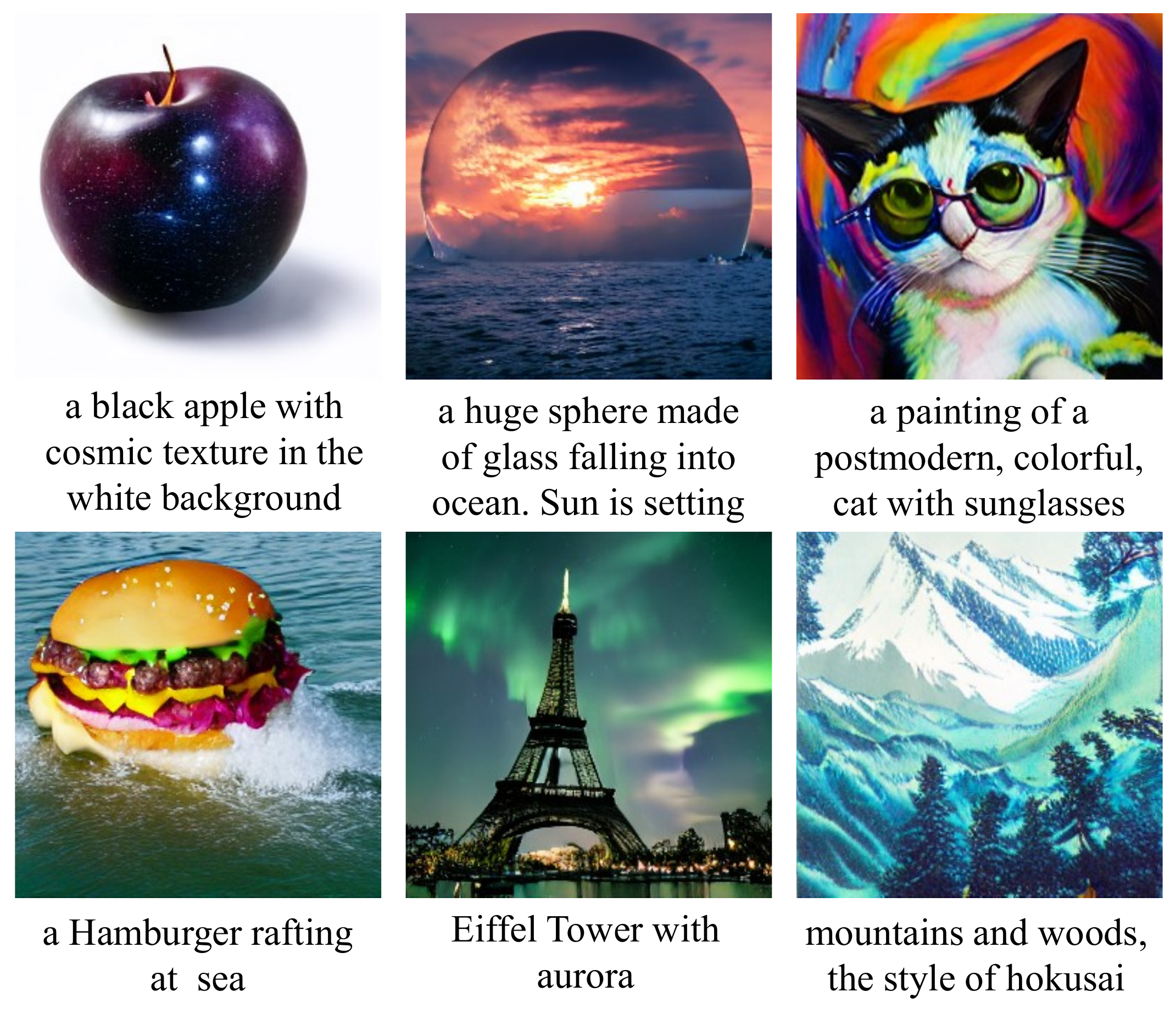}
    \caption{Selected 256$\times$256 samples generated by our model. Lformer can generate both photorealistic and artistic content.}
    \label{intro_show}
\end{figure}

Despite the success these methods have achieved, they still suffer from the slow generation problem. Most of the existing generative transformers~\cite{DALLE, Cogview1, NUWA, Make_A_Scene, Parti} leverage autoregressive~(AR) models, which rearrange the 2D image tokens in the raster-scan order and decode only one token in each step. Therefore, they take 1024 forward steps with a large-scale transformer to generate $32\times32$ tokens for a single $256\times256$ image, which consumes about 30 seconds on a V100 GPU. However, the users have a high requirement for generation speed, since they usually need to generate dozens of candidates to obtain the image they expect.

\begin{figure}[tbp]
\centering
\includegraphics[width=0.95\columnwidth]{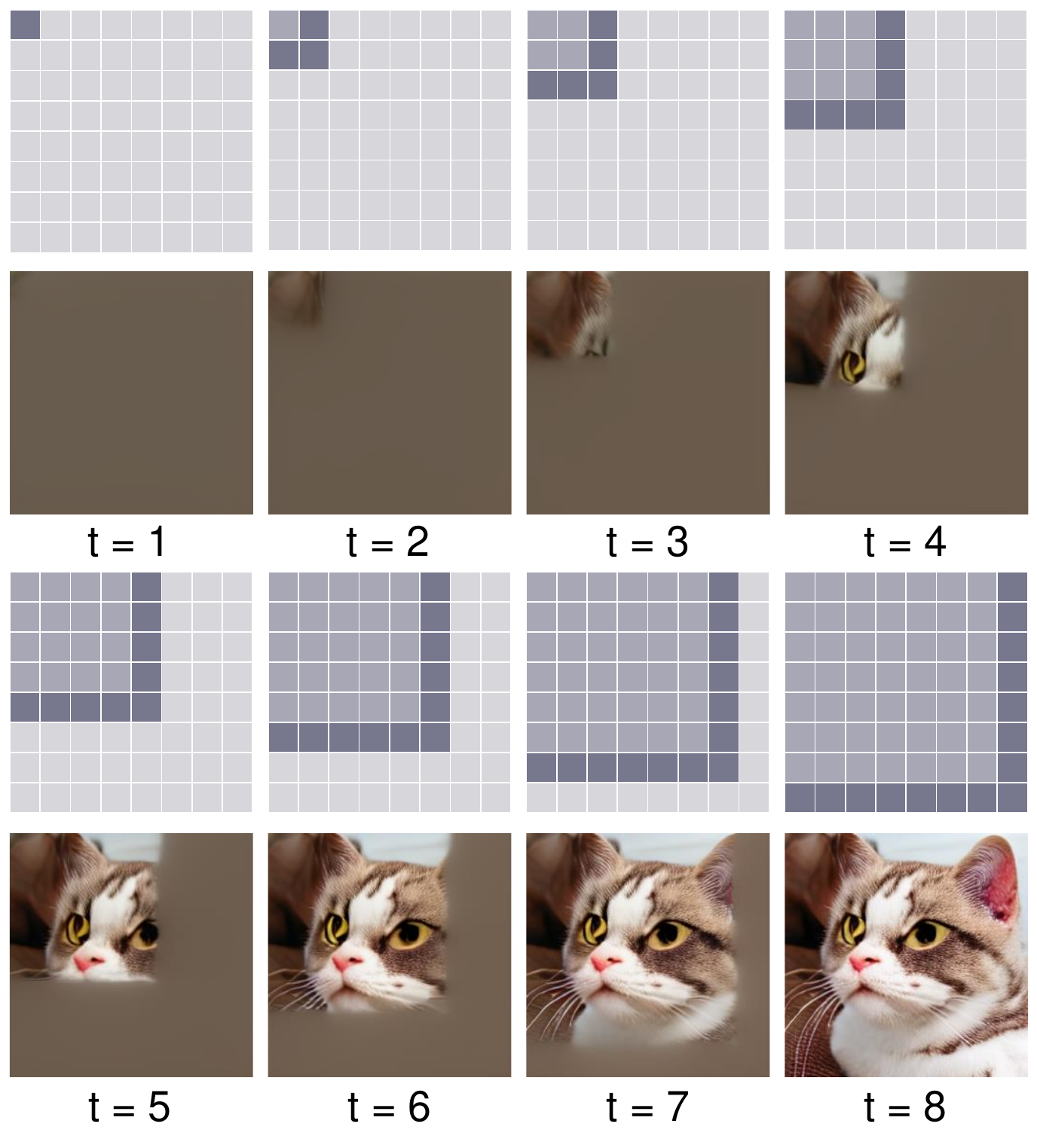}
\caption{Generation process of Lformer~(taking $h{\times}w=8\times8$ as an example). We divide an image token map into $h$ mirrored L-shape blocks from the top left to the bottom right. At each step, we generate all the tokens in an L-shape block parallelly.}
\label{intro}
\end{figure}

Some recent works~\cite{MaskGIT, UFC-BERT, cogview2, VQDIFF} can generate $32\times32$ tokens within a few steps with the design of parallel decoding strategy like Mask-Predict~\cite{MaskPredict}) and Discrete Diffusion~(DD)~\cite{VQDIFF, DD2}, which has greatly improved the inference speed. However, the speed of a transformer depends not only on the number of decoding steps but also on the forward time of each step, which is quadratic with the sequence length. The existing methods~\cite{MaskPredict, UFC-BERT, cogview2, MaskGIT} forward with full $h{\times}w$ tokens in every step. Thus there is still much room for the optimization of the inference time at each step.

To better accelerate the generative transformers while keeping good generation quality, we propose Lformer, a semi-autoregressive~(SAR) T2I generation model that parallelly decodes image tokens in L-shape blocks, which can generate high-quality images at a high speed. Lformer is fast because it generates images within a small number of steps and processes only partial tokens in each step.
Specifically, as shown in \cref{intro}, we generate image tokens in a $h{\times}h$ window and divide these tokens into $h$ mirrored L-shape blocks from the top left to the bottom right. At each step, we generate the tokens in a block parallelly. The newly generated tokens expand the context to a larger square, and we finish the generation in $h$ steps. 

Lformer achieves parallelism by leveraging the 2D structure of image tokens. Thus it can keep good generation quality while greatly improving the sampling speed. 
The area predicted by Lformer in each step is adjacent to the square context. Such adjacent areas have a stronger association with the context than the non-adjacent area, which will make the operation of predicting them easier. The process of extending adjacent tokens is similar to the process of autoregressive models and our model can benefit from the techniques of unidirectional models, such as teacher forcing~\cite{TeacherForcing} for efficient training and attention cache~\cite{Fastseq} for fast inference. Note that only Lformer achieves such compatibility between parallelism and unidirectional techniques in T2I generation due to its semi-autoregressive design, whereas NAR methods can not leverage the techniques like attention cache.

Additionally, the pretrained Lformer has the potential capability to edit images without the requirement for additional fine-tuning. This is because the discrete tokens have strong space mapping relations with the generated images, and our generation process is trackable. So when encountering a local defect, we can roll back to the early steps for regeneration or edit the image with a bounding box and a text prompt, as shown in ~\cref{sec:edit}.

We train our model on three T2I datasets of different scales: Multi-Modal CelebA-HQ~\cite{MMCelebA} containing 30K face images, CC3M~\cite{CC} and a subset of LAION~\cite{LAION400} respectively containing 3M and 90M general domain images for large-scale training. Compared with the existing transformer-based methods, our method is much faster while achieving even better a Fréchet Inception Distance~(FID)\cite{FID} score on MMCelebA-HQ. Lformer also shows competitive performance on MS-COCO 2014 in both zero-shot and fine-tuning settings.  Moreover, our pretrained model can even generate counterfactual images with good quality, as shown in \cref{intro}.

\section{Related Work}
\subsection{Text-to-Image Generation} The task of text-to-image~(T2I) generation aims to synthesize images from the given text descriptions. The early deep-learning-based T2I models like AttnGAN~\cite{AttnGAN}, DM-GAN~\cite{DM-GAN}, DF-GAN~\cite{DF-GAN} build upon Conditional Generative Adversarial Nets~(CGAN)~\cite{cGAN}. These models can generate high-quality images for specific domains like birds~\cite{CUB} and flowers~\cite{Oxford}, but the general-domain generation presents great challenges for them. Later, a series of works~\cite{VQVAE2, DALLE, Cogview1, NUWA, Make_A_Scene, RQVAE, Parti, Imagebart} adopt a two-stage framework that leverages an image tokenizer to compress an image to a sequence of discrete tokens and treat the T2I generation as a sequence-to-sequence modeling problem. They train a large-scale transformer\cite{Transformer} with a hundreds-million-scale dataset and achieve great advancement on the general-domain generation. Recently, diffusion models like GLIDE~\cite{GLIDE}, DALL·E 2~\cite{DALLE2}, Imagen~\cite{Imagen}, LDM~\cite{LDM} also show great effectiveness on general-domain T2I generation.

\subsection{Autoregressive Models for T2I}
The early works of Autoregressive~(AR) Models~\cite{pixelcnn, pixelcnn++, pixelsnail, iGPT} directly generate at the pixel level, whose resolution is limited by heavy computation. Later works like DALL·E~\cite{DALLE} and Cogview~\cite{Cogview1} compress an image with VQ-VAE~\cite{VQVAE} and model the text and image tokens with a transformer. Esser~\etal propose VQGAN~\cite{VQGAN}, an image tokenizer that can reconstruct images with better photorealism. N\"UWA~\cite{NUWA} proposes a 3D transformer encoder-decoder framework to further leverage videos. Make-A-Scene~\cite{Make_A_Scene} introduces additional controlling elements, such as the segmentation map. Parti~\cite{Parti} scale the transformer up to 20B parameters, achieving an impressive zero-shot FID score~\cite{FID} of 7.23 on MS-COCO 2014~\cite{MSCOCO}.

\subsection{Parallel Decoding Models for T2I}
The current transformer-based T2I models that can achieve parallel decoding are mainly 
non-autoregressive (NAR) Models. They usually achieve faster inference speed than autoregressive models while keeping similar generation quality. Some existing NAR methods~\cite{MaskGIT, cogview2} for T2I are inspired by the works of natural language generation~(Mask-Predict~\cite{MaskPredict}, GLAT~\cite{GLAT}, GLM~\cite{GLM}). UFC-BERT~\cite{UFC-BERT} and MaskGIT~\cite{MaskGIT} adopt the bidirectional mask-and-predict framework like BERT~\cite{BERT} and Mask-Predict~\cite{MaskPredict}. The key idea of these methods is to predict all tokens simultaneously in parallel but only replace the mask tokens that get the highest confidence in each step. As for other NAR methods, Cogview2~\cite{cogview2} combines autoregressive blank filling with hierarchical generation. There are also works like VQ-Diffusion~\cite{VQDIFF} that leverage diffusion in the token space to achieve parallelism.   

\section{Method}
\subsection{Preliminaries}
Most of autoregressive~(AR) and non-autoregressive (NAR) methods are based on two-stage image generation frameworks~\cite{VQGAN, DALLE, Cogview1}, and the key difference between them lies in the decoding process in the second stage. Lformer is also carefully designed for a more efficient decoding scheme. Therefore, before introducing the details of Lformer, we first review the general framework of two-stage image generation methods.

In the first stage, the algorithm learns an encoder E, a decoder D, and a codebook $e=\{e_k\}^{K}_{k=1}\subset{\mathbb{R}^{n_e}}$, where $n_e$ is the dimensionality of codes, $K$ is the size of codebook. They work together to compress an image $x\in{\mathbb{R}}^{H{\times}W{\times}3}$ into discrete tokens $s=\{s_i\}^N_{i=1}, s_i{\in}\{0, ..., |e|-1\}$ and reconstruct the image with these tokens. Specifically, the encoder E encodes an image into spatial codes $\hat{e}=\mathrm{E}(x)\in\mathbb{R}^{h{\times}w{\times}{n_e}}, h=H/f, w=W/f, f=2^m$, where $f$ is the downsampling factor, $m$ is the number of downsampling blocks in the encoder E. This $h{\times}w{\times}{n_e}$ tensor can be viewed as $h{\times}w$ codes of $\hat{e}_{ij}\in\mathbb{R}^{n_e}$, where $i, j $ denote the code indices in $h{\times}w$. Each spatial code $\hat{e}_{ij}$ in $\hat{e}$ is quantized by replacing it with its nearest prototype vector in the codebook:
\begin{equation}
    \mathbf{q}(\hat{e}_{ij})=e_k, 
    \text{where } k=\mathop{\arg\min}\limits_{k}||\hat{e}_{ij}-e_k||
\end{equation}
where $\mathbf{q}$ denotes the quantization operation. We can reconstruct the image from the quantized codes $\tilde{e}$ = $\mathbf{q}(\hat{e})\in\mathbb{R}^{h{\times}w{\times}{n_e}}$ by:
\begin{equation}
    \hat{x} = \mathrm{D}(\tilde{e}) = \mathrm{D}(\mathbf{q}(\mathrm{E}(x)))
\end{equation}
The above process is supervised by reconstruction loss~\cite{VQVAE, VQVAE2}, optionally with perceptual loss~\cite{Perceptual, VQGAN} and GAN loss~\cite{VQGAN}.
The quantized codes $\tilde{e}$ can be further represented by their indices in the codebook, which we call image tokens:
\begin{equation}
\begin{split}
    s =\{s_{ij}\}\in\{1, ..., K\}^{h{\times}w}\\
    s_{ij} = k \text{ such that } \tilde{e}_{ij}=e_k    
\end{split}
\end{equation}
By mapping the image tokens $s$ back to their corresponding codebook entries, we can recover the quantized codes ${\tilde{e}}$ and further decode it to an image by $\hat{x} = \mathrm{D}(\tilde{e})$.

In the second stage, the key is to learn the prior $p(s|c)$ generation model,
where $c$ denotes the condition. Take AR methods as example, they generally rearrange $s^{h{\times}w}$ into raster-scan order $s^{h{\cdot}w}$ and predict one token at each step:
\begin{equation}
    p(s|c) = \prod_{t=1}^{h{\cdot}w} p(s_{t}|s_{[1:t-1]}, c)
\end{equation}
where $s_{[1:t-1]}$ denotes $\{s_{1}, s_{2}, ..., s_{t-1}\}$. Once a sequence $s$ is sampled, we can obtain the corresponding generated image by decoding $s$ with the decoder $\mathrm{D}$.

\begin{figure}[tbp]
\centering
\begin{subfigure}{0.4\columnwidth}
\includegraphics[width=\columnwidth]{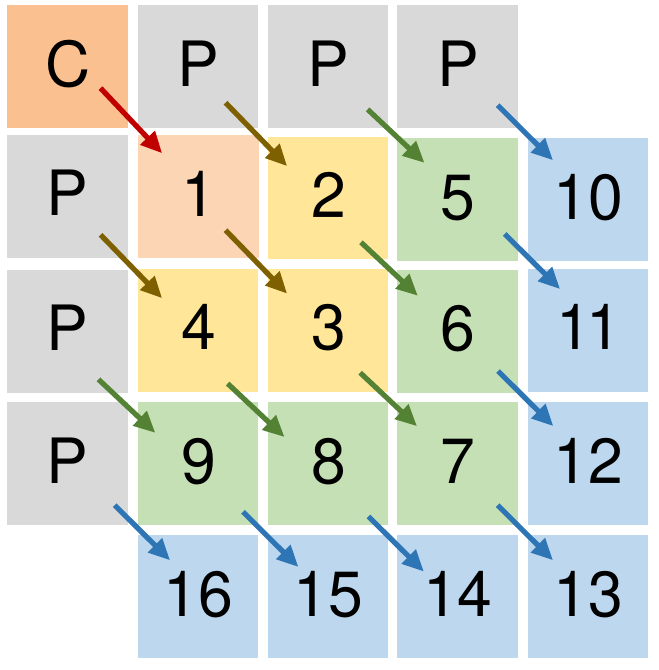}
\caption{L-order}
\label{infer}
\end{subfigure}
\hfill
\begin{subfigure}{0.4\columnwidth}
\includegraphics[width=\columnwidth]{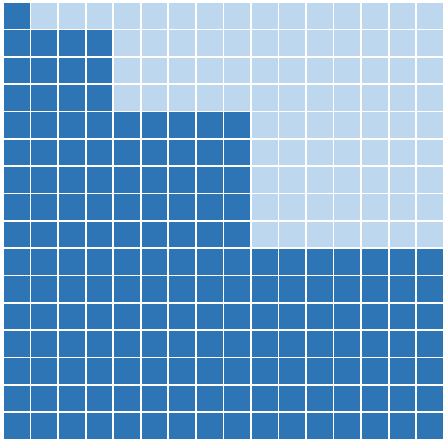}
\caption{Attention Mask}
\label{attention}
\end{subfigure}
\hfill
\begin{subfigure}{\columnwidth}
\includegraphics[width=\columnwidth]{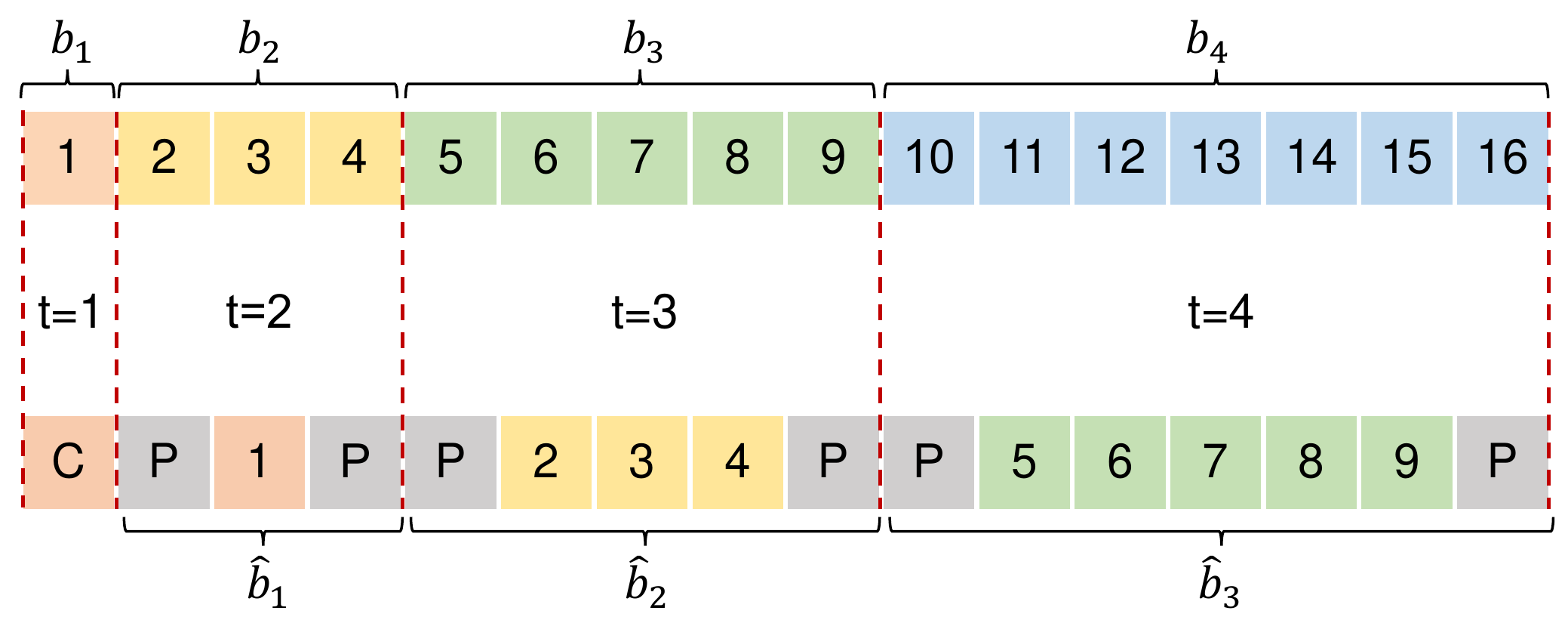}
\caption{Alignment for transformer}
\label{align}
\hfill
\end{subfigure}
\caption{L-shape Parallel Decoding. (a) We rearrange the tokens in L-order and  generate the tokens in an L-shape Block~(marked in colors) in a step parallelly. $\mathrm{C}$ denotes the condition, and $\mathrm{P}$ denotes a pad token. (b) Causal attention mask for the transformer implementation of Lformer. The tokens in a block can attend to the context of previously generated tokens which forms a square. (c) We add pad tokens at both sides of the previous L-block to align with the next L-block for transformer implementation.}
\end{figure}

\subsection{L-shape Parallel Decoding}
Although AR and NAR methods have achieved great success, the AR methods suffer from low generation speed when achieving high generation quality, and the NAR methods suffer from unstable generation quality when achieving high generation speed.
In order to accelerate the generative transformers while keeping good generation quality, we propose a semi-autoregressive decoding strategy, L-shape Parallel Decoding.
It can parallelly decode tokens in a step while considering the unidirectional characteristic of AR models by leveraging the 2D structure of image tokens. 
Specifically, we generate square images~(e.g. $256\times256$), and we have $h=w$ for image tokens. We divide $h{\times}h$ image tokens into $h$ blocks from the top left to the bottom right, with each block in a mirrored L-shape~(the first block only has the single top left token as a special case) numbered from 1 to $h$. As shown in \cref{infer}, we rearrange the tokens in the order that considers the block order first and iterate clockwise within a block. So the $t$-th L-block will contain tokens of $s_{[(t-1)^2+1:t^2]}$ and the context of previous t steps is a square $s_{[1:(t-1)^2]}$.
We decode $h{\times}h$ tokens in $h$ steps and parallelly decode all the tokens in the $t$-th block at time step $t$, with $t=1, ..., h$:
\begin{equation}
    p(s|c) = p(s_1|c)\cdot\prod_{t=2}^{h} 
    p(s_{[(t-1)^2+1:t^2]}|s_{[1:(t-1)^2]}, c)
    \label{parallel}
\end{equation}
where $p(s_1|c)$ means we directly generate the token $s_1$ from the condition $c$ for $t=1$. For clarity and simplicity, we denote the $t$-th L-block $s_{[(t-1)^2+1:t^2]}$ as $b_t$, and the square context $s_{[1:t^2]}$ as $s_{<t^2}$ in the rest of this paper.

To implement \cref{parallel} with a transformer-based generation model, we need to fill the sequence with pad tokens for alignment, since the output length for a transformer is the same as its input length. Each L-block $b_t$ contains $2t-1$ tokens, and the number of tokens contained in adjacent blocks differs by two. So we insert pad tokens at both sides of the $(t-1)$-th L-block $b_{t-1}$ to align with the $t$-th L-block $b_{t}$, as shown in \cref{align}. The alignment for the transformer can be formulated as:
\begin{align}
    &\hat{b}_t = \text{pad}(b_t) =[ {\text{P}}; b_t ; {\text{P}}]\\
    &[b_1'; b_2', ...; b_t'] = \text{Transformer}([c; \hat{b}_1; \hat{b}_2, ...; \hat{b}_{t-1}]))
\end{align}
where $b_t$ is the $t$-th L-block of ground truth tokens, ${\text{P}}$ denotes a pad token, [;] denotes the concatenation, $\hat{b}_t$ denotes the padded $b_t$. 
The model's output at the place corresponding to the $\hat{b}_{t-1}$ will be used to predict $b_t$, and we denote the model's prediction for $b_t$ as $b_t'$. We train the model with the average cross-entropy~\cite{VQGAN, Parti} of all $h{\times}h$ tokens: $L=\frac{1}{h^2}\sum_{t=1}^{h}{\text{CE}(b_t', b_t)}$, where CE denotes the sum of the cross-entropy of tokens in $b_t'$ and $b_t$. 

With such a design, we can train Lformer in the way of teacher-forcing~\cite{TeacherForcing} with a causal attention mask shown in \cref{attention}. This is efficient for training since all the tokens contribute to the cross-entropy loss. Besides, at inference time, we can further accelerate with the attention cache mechanism~\cite{Fastseq}, by which we cache the keys and values of previous steps to avoid redundant calculations.

\begin{figure}[tbp]
\centering
\includegraphics[width=\columnwidth]{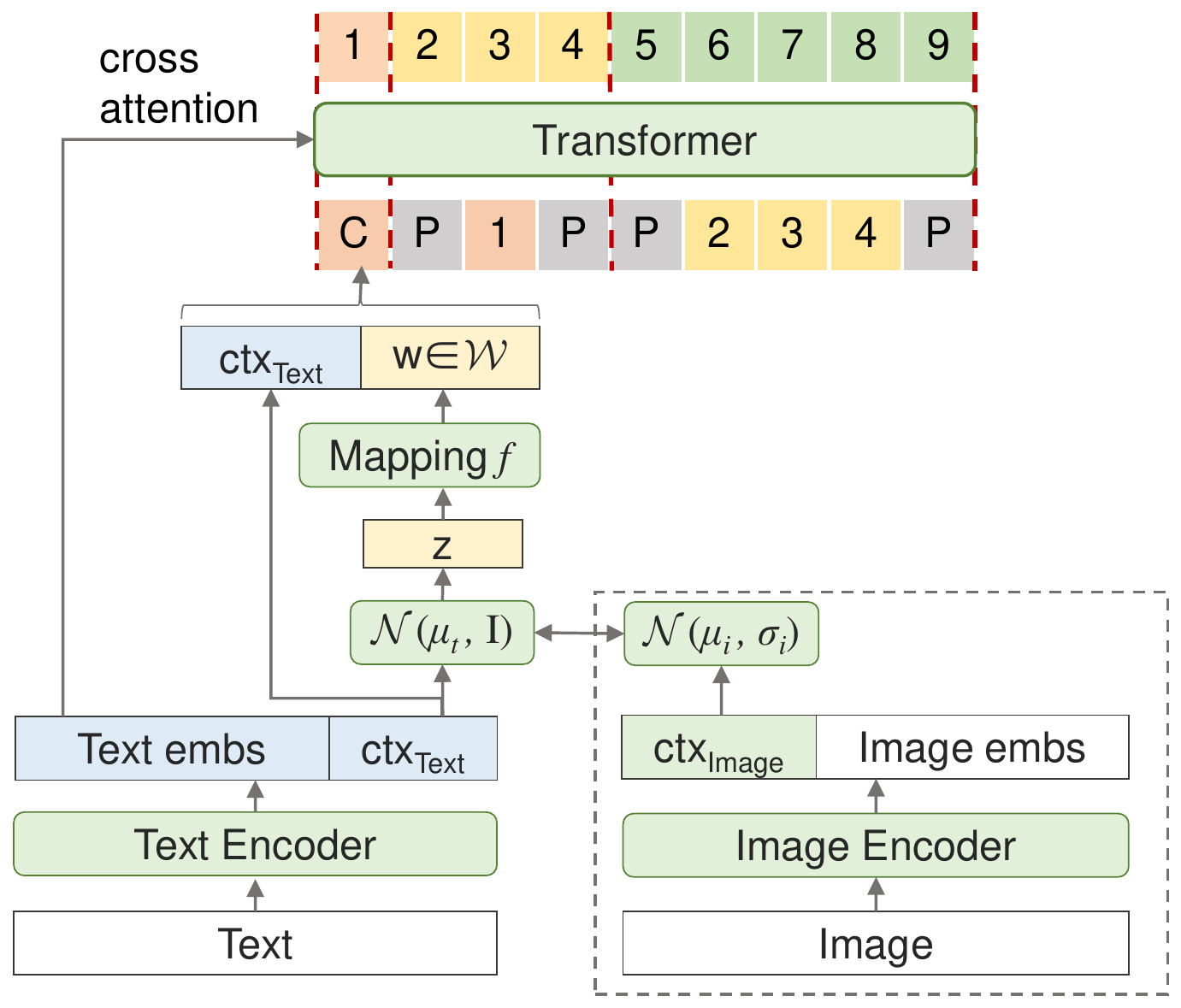}
\caption{Model structure of Lformer. We adopt CVAE to introduce a latent variable $w$ for global representation to the inconsistency of parallel generation. We concatenate $w$ to the global text feature as the final condition. The part in the dashed box is removed at inference time.}
\label{structure}
\end{figure}

\subsection{Global Representation for Parallel Decoding}
Since Lformer decodes the $2t-1$ tokens in the L-block parallelly, the probabilities of tokens in the same block are independent, which will make the model partially exhibit \textit{conditional independence}~(discussed in Sec 2.3 of \cite{NAR}, the multimodality problem of parallel decoding). To preserve the stability of the generation process, an extra global representation for the constraint is required in Lformer.
Similar to~\cite{VAERNN, VT}, we adopt the Conditional Variational Auto-Encoder~(CVAE)~\cite{CVAE} to enhance the generation consistency.
\cref{structure} shows the model structure of Lformer. We use CLIP~\cite{CLIP} text encoder to extract text features, treat the CLIP image encoder $\phi$ as the encoder of CVAE, and treat the transformer model $\theta$ as the decoder of CVAE, then we have:
\begin{equation}
    \begin{split}
        \mathcal{L}_{\text{CVAE}}(x,\hat{c};\theta, \phi) = &-\text{KL}(q_\phi(z|x,\hat{c})||p_\theta(z|\hat{c})))\\
        &+\mathbb{E}_{q_\phi(z|x,\hat{c})}({\text{log }p_\theta(x|z,\hat{c})})
    \end{split} 
    \label{loss}
\end{equation}
where $x$ is an image sample, $\hat{c}$ is the text context embedding extracted by CLIP text encoder, KL denotes Kullback-Leibler~(KL) Divergence, $q_\phi(z|x,\hat{c})$ is a Gaussian distribution $\mathcal{N}(\mu_i, \sigma_i)$ estimated by the CLIP image encoder $\phi$, $p_\theta(z|\hat{c})$ is a Gaussian distribution $\mathcal{N}(\mu_t, \text{I})$ estimated by the CLIP text encoder. 
The condition $c$ for the transformer input is:
\begin{equation}
        {c} = [\hat{c}; f(z)]
\end{equation}
where $f(\cdot)$ is a mapping function consisting of fully connected layers that map the Gaussian latent variable z to the $\mathcal{W}$ space~\cite{stylegan}.

In the framework of CVAE, we estimate a Gaussian distribution for each image and constraint this distribution to the distribution estimated by its corresponding text with KL Divergence. With such a design, we can sample a latent variable $z$ from $p_\theta(z|\hat{c})$ as the representation of the image instance we sample for the given text $\hat{c}$.
The second term of \cref{loss} is the reconstruction loss, which we can estimate by \cref{parallel} with cross-entropy since we have $x=\mathrm{D}(s)$. 

Note that we do not need the CLIP image encoder $\phi$~(the part in the dashed box) at inference time since we can sample $z$ from $p_\theta(z|\hat{c})$. Our final model  consists of a transformer, a mapping function $f$, and a CLIP text encoder. We also use cross-attention to enhance the text semantics. 

\section{Computation Complexity Analysis}
Lformer is faster not only for a small number of decoding steps but also for its context structure. The computation complexity of a transformer mainly comes from the attention mechanism~\cite{Transformer}. We denote the length of the context as $N$~($h{\times}h$, usually 256 or 1024), the hidden dimension as $D$, and the inference steps as $T$. The computation in the attention mainly comes from the matrix multiplication of $Q{\cdot}K^\mathrm{T}{\cdot}V$, where $Q, K, V\in\mathbb{R}^{N{\times}D}$. The total number of multiplications of a bidirectional model~(e.g., the BERT model used in NAR methods) is proportional to $TN^2D$. As for Lformer, while working with the attention cache, we have Q$\in\mathbb{R}^{(2t-1){\times}D}$ and K, V $\in\mathbb{R}^{t^2{\times}D}$.
Thus the number of multiplication operations is 
\begin{equation}
    M=\sum_{t=1}^{\sqrt{N}}(2t-1)t^2D=(\frac{1}{2}N^2+\frac{2}{3}N^\frac{3}{2}-\frac{1}{6}\sqrt{N})D
\end{equation}
When we have N=1024, T=$\sqrt{N}$, D=1024, the number of attention multiplication $M$ for a bidirectional model is 33B whereas Lformer gets its $M$ of 0.5B, which is much smaller.

\section{Experiments}

\subsection{Experimental Setup}
\noindent\textbf{Datasets}
We train Lformer on three datasets with different scales: Multi-Modal CelebA-HQ~\cite{MMCelebA}~(MMCelebA-HQ) containing 30 thousand samples, Conceptual Captions~(CC3M)~\cite{CC} containing 3 million samples, and a subset of LAION400M~\cite{LAION400} containing 90 million images, which we call LAION90M. 

\noindent\textbf{Implementation Details}
We use CLIP-ViT-B/16~\cite{CLIP} as the text encoder and freeze the parameters during training. We generate 256$\times$256 images for all datasets. To save GPU memory, we adopt techniques of the ZeRO optimization~\cite{ZeRO}, block sparse attention~\cite{SparseATT}, and activation checkpointing. All models are trained in float16 precision, and we adopt PB-relax and Sandwich LaryerNorm proposed by~\cite{Cogview1} to keep stability. We use top-$k$~\cite{VQGAN, Cogview1, RQVAE} and top-$p$~\cite{Nucleus_Sampling, RQVAE} sampling during inference. 
The structures of the transformers vary by dataset, and we show their details in the corresponding sections. In the rest of the paper, we use L, H, and D to denote the number of layers, attention heads, and the hidden dimension of a transformer, respectively.
More details about the datasets and implementation are available in the Appendix. 

\subsection{Comparison of Inference Speed} 
We first verify the speed of Lformer and compare it with the existing transformer-based methods in~\cref{latency}. We generate a batch of 32 images with $32\times32$ tokens, and the decoding time of the VQGAN is included in the ``BatchTime''. All the tests are done with the same machine with a V100 GPU, and we report the average time of 10 batches.
We follow the setting of VQ-DIFFUSION~\cite{VQDIFF} and test with a transformer of L=19, H=16, and D=1024 except for RQVAE~\cite{RQVAE}. RQVAE generates images with a token map of $8\times8\times4$, and it has an extra Depth Transformer. Thus we separately compare an RQVAE of L=26+4~(4 layers for Depth Transformer.), H=20, D=1280, with an Lformer of L=30, H=20, D=1280 by generating $16\times16$ tokens. VQ-Diffusion~\cite{VQDIFF} can generate in 25 to 100 steps, and we test it with 32 steps for a fair comparison. 

As shown in \cref{latency}, Lformer achieves the fastest inference speed of 3.57s for a batch when working with the attention cache~\cite{Fastseq}. It also surpasses RQVAE~\cite{RQVAE} with 2.37s on the $16{\times}16$ tokens generation. 
Note that only the unidirectional models like AR and Lformer can leverage attention cache.
We also find that the AR model~(Our implemented version of AR uses cross-attention) is surprisingly fast with the attention cache. This is because the early steps of a unidirectional model have a partial context~~($<h{\times}h$ tokens) while a NAR model forwards with a full context~($h{\times}h$ tokens) at each step.

\begin{table}[t]
\centering
\setlength{\tabcolsep}{2pt}
\begin{tabular}{lccccc}
\toprule
Methods                    &Steps   &Params &\makecell{Batch\\Time(s)}    &s/img   &img/s\\
\hline
AR~\cite{VQGAN}            &1024    &322M   &1270.0       &39.7      &0.025\\
AR-cache~\cite{VQGAN}      &1024    &322M   &33.94        &1.06      &0.942\\
VQ-Diffusion~\cite{VQDIFF} &32      &370M   &86.02        &2.67      &0.372\\
MASKGIT~\cite{MaskGIT}     &8       &322M   &21.05        &0.672     &1.49\\
Lformer~(Ours)             &32      &322M   &26.0         &0.813     &1.23\\
Lformer-cache~(Ours)       &32      &322M   &\textbf{3.57}  &\textbf{0.112}     &\textbf{8.96}\\
\hline
RQVAE~\cite{RQVAE}         &256     &654M   &5.32         &0.166     &6.02\\
Lformer-cache~(Ours)       &16      &798M   &\textbf{2.37}  &\textbf{0.074}     &\textbf{13.5}\\
\bottomrule
\end{tabular}
\caption{Speed comparison. We compare to the existing transformer-based methods with the same structure of L=19, H=16, and D=1024. We generate a batch of 32 images with $32\times32$ tokens except for RQVAE~\cite{RQVAE}. We separately compare with RQVAE using a fair setting. Note that only unidirectional models like AR and Lformer can leverage attention cache.}
\label{latency}
\end{table}

\begin{table}[t]
\centering
\setlength{\tabcolsep}{5pt}
\begin{tabular}{lcccccc}
\toprule
Model                   &Type   &Params  &FID$\downarrow$  &s/img$\downarrow$\\
\hline
VQGAN$^*$~\cite{VQGAN}          &AR   &406M    &19.52     &3.676\\
UFC-BERT$^*$~\cite{UFC-BERT} &NAR  &406M       &23.62 &0.168\\
LPD                             &SAR  &406M     &20.93    &0.046\\
\hline
UFC-BERT+CVAE$^*$               &NAR  &406M     &22.65 &0.168\\
LPD+CVAE                        &SAR  &406M     &19.91   &0.046\\
\hline
Lformer-S      &SAR      &102M     &23.30  &0.030\\
Lformer-M      &SAR      &229M     &22.13  &0.036\\
Lformer-L      &SAR      &406M     &19.91  &0.046\\
Lformer-E      &SAR      &1B       &18.60  &0.076\\
\bottomrule
\end{tabular}
\caption{Comparison on MMCelebA-HQ. We test the effect of decoding strategies, CVAE module, and model scales. LPD+CVAE is equivalent to Lformer-L. $^* $ denotes our implementation.}
\label{tab:celeba}
\end{table}

\subsection{Comparison on MMCelebA-HQ}
\begin{figure}[tbp]
\centering
\includegraphics[width=0.96\columnwidth]{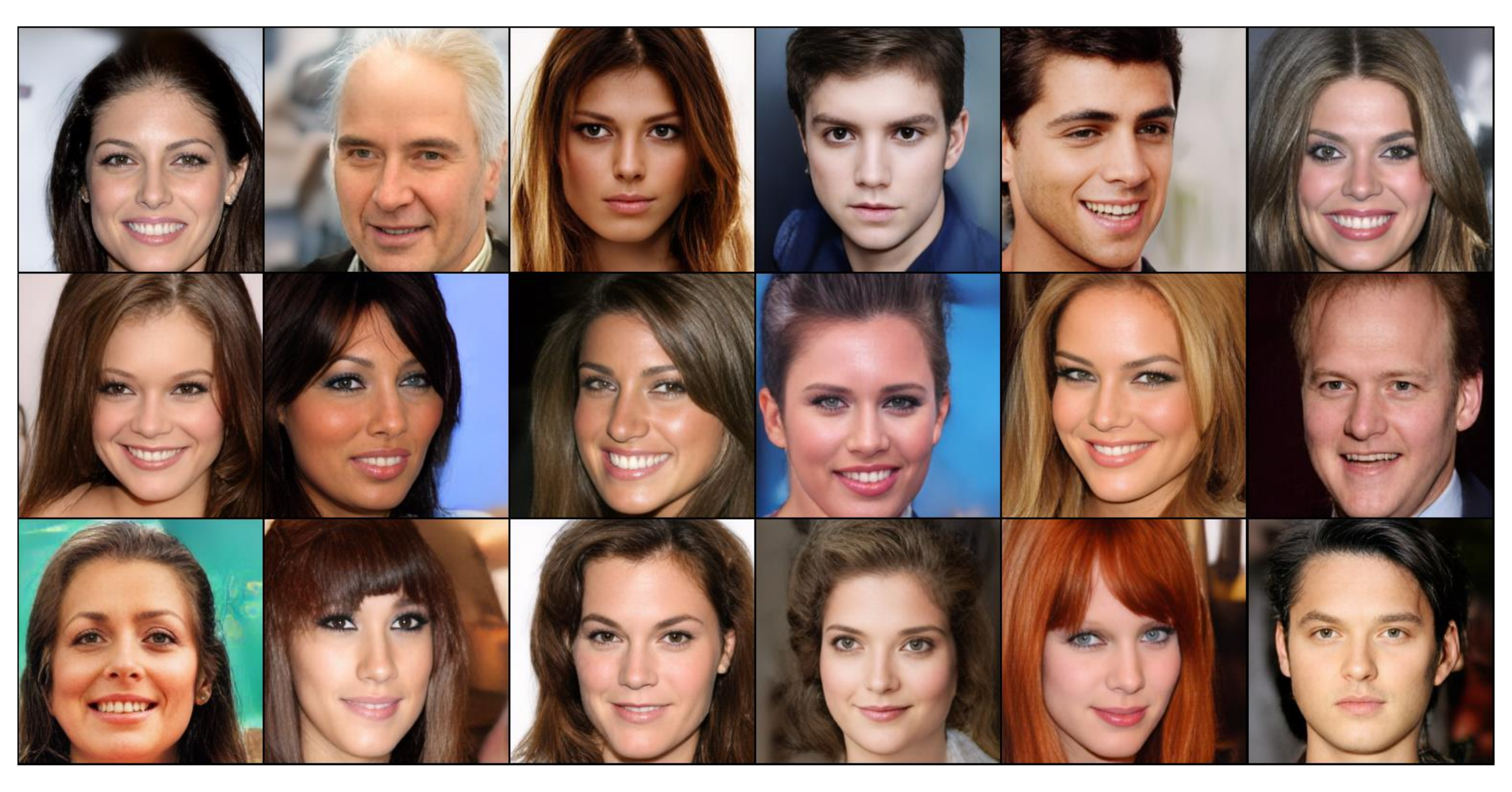}
\caption{Random samples of 256$\times$256 generated by Lformer-L with the text prompts from the MMCelebA-HQ validation set.}
\label{faces}
\end{figure}

After verifying the efficiency of Lformer, we further test its generation quality on MMCelebA-HQ~\cite{MMCelebA}. Following UFC-BERT~\cite{UFC-BERT}, we set the transformer structure to L=24, H=16, and D=1024 and generate in the latent of 16$\times$16 tokens. 
As shown in ~\cref{faces}, Lformer can generate high-quality images with the text prompts from the MMCelebA-HQ dataset.
To demonstrate the effectiveness and scalability of our method, we further compare models with different decoding strategies, module settings, and model scales by their Fréchet Inception Distance~(FID)\cite{FID} scores and generation speed. The implementation details of these models are available in the Appendix.

\noindent\textbf{Strategy Comparison.}\quad
We compare models with decoding strategies of the standard autoregressive~(AR) scheme (e.g., VQGAN~\cite{VQGAN}), the mask and predict~(NAR) scheme(e.g., UFC-BERT~\cite{UFC-BERT}), and our L-shape Parallel Decoding~(LPD). During the comparison, we do not use any global representation and use the same first-stage model and transformer model.
As shown in \cref{tab:celeba}, the AR method has better generation quality but generates slowly, and the NAR method shows a great improvement in generation speed but obtains a higher FID score of 23.62. By comparison, our LPD obtains a lower FID score of 20.93 with a much faster speed of 0.046 s/img. This shows that Lformer can keep good generation quality with a high inference speed since it leverages the prior of the 2D structure of images.

\noindent\textbf{Module Effect.}\quad
We compare the model with and without the global representation provided by CVAE to 
show its effect. We also apply it to the NAR method like UFC-BERT which also uses parallel decoding. Note that LPD+CVAE is equivalent to the full model Lformer-L. The model with the global representation achieves a lower FID score of 19.91 because the global representation relieves the conditional independence problem in parallel decoding. 

\noindent\textbf{Scalability.}\quad 
Generally, the performance is gradually improved as the model size grows. Besides, we also observe quality improvement when scaling the model from 229M to 406M. The phenomenon of quality improvement is also observed in Parti\cite{Parti}.


\begin{table*}[t]
\centering
\setlength{\tabcolsep}{5pt}
\begin{tabular}{lcccccccc}
\toprule
\quad  &\multicolumn{3}{c}{Zero-shot} &Finetuned &\multicolumn{3}{c}{\quad}\\
\cmidrule(lr){2-4} \cmidrule(lr){5-5}
{Models} &FID$\downarrow$    &IS$\uparrow$    &CLIPSIM$\uparrow$  &FID$\downarrow$ &Type    &Params  &Data      &Computation\\
\hline
VQ-Diffusion$^*$~\cite{VQDIFF} &32.8$^*$   &15.2$^*$  &0.2539$^*$   &-  &DD   &0.5B  &12M      &-\\
DALL·E~\cite{DALLE}            &27.5   &17.9  &-        &-  &AR   &12B    &250M     &430K$\times$6K\\
Cogview~\cite{Cogview1}        &27.1   &18.2  &\textbf{0.3325} &- &AR  &4B   &30M   &144K$\times$6K\\
RQVAE~\cite{RQVAE}             &16.9   &23.76$^*$ &0.3058$^*$   &-     &AR  &3.9B   &30M      &-\\
N\"UWA~\cite{NUWA}             &-      &-     &-   &12.9  &AR   &0.87B   &2.9M+Video &50M$\times$128\\
CogView2~\cite{cogview2}       &24.0   &22.4  &-   &17.4  &NAR &6B     &30M      &300K$\times$4K\\
Lformer-E~(CC3M)                &24.11	 &19.78	&0.3052	  &-  &SAR  &1B	  &2.5M      &240K$\times$1K\\
Lformer-E~(LAION90M)          &\textbf{15.01}  &\textbf{26.22} &0.3289   &\textbf{9.57}  &SAR  &1B	  &90M       &300K$\times$1K\\
\hline
ERNIE-ViLG~\cite{ERNIE-ViLG}     &14.7   &-     &-   &7.9   &AR   &10B   &145M     &-\\
Make-A-Scene~\cite{Make_A_Scene} &11.84  &-     &-   &7.55  &AR   &4B    &35M+SEG   &170K$\times$1K\\
Parti~\cite{Parti}               &\textbf{7.23} &-   &- &\textbf{3.2} &AR  &20B    &1.8B+4B   &450K$\times$8K\\
\bottomrule
\end{tabular}
\caption{Performance comparison on MS-COCO~2014~\cite{MSCOCO}. Lformer samples 16 images per text prompt and reranks with a CLIP~\cite{CLIP} model. $*$ denotes the result that we evaluate with their released model~(also rerank with 16 samples). $\downarrow$ denotes lower better and $\uparrow$ denotes higher better. The `Computation' column shows the training cost in the form of `training steps {$\times$} batch size'. }
\label{zero-shot}
\end{table*}

\begin{figure*}[tbp]
    \centering
    \includegraphics[width=0.95\textwidth]{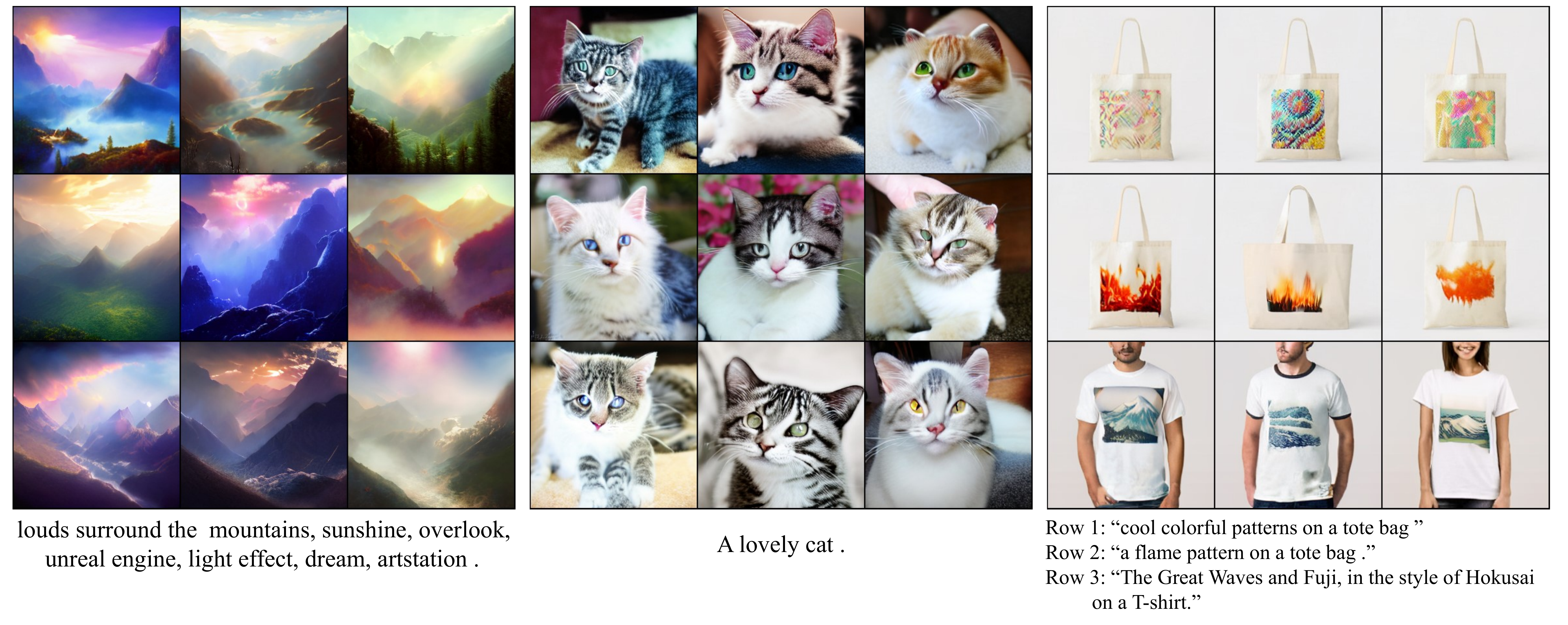}
    \caption{Diversity and composition capability. We show the images generated with the same text prompt. We also try the combination of a pattern and a carrier~(tote bag or T-shirt) to show the creativity of the model.}
    \label{diversty}
\end{figure*}

\subsection{Comparison on MS-COCO}
The current text-to-image research~\cite{Cogview1, cogview2, Make_A_Scene, DALLE, Parti,ERNIE-ViLG} mainly evaluates their models on MS-COCO 2014, where they pretrain a model on a large-scale dataset and then measure both zero-shot and fine-tuned generation quality on MS-COCO. We follow this setting and train Lformer models on CC3M and LAION90M, respectively. To fit such large datasets, we also scale up the model to 1B parameters with the transformer structure of L=24, H=16, and D=1536. We adopt the Transformer Classifier-Free Guidance~(CFG) \cite{Make_A_Scene} with a scale of 3.0 and generate with $32\times32$ tokens. For each caption, we generate 16 images and select the best one with CLIP reranking score~\cite{DALLE}. This model takes about 9 seconds~(with CFG) to generate a batch of 16 images on a V100 GPU.

Following \cite{DALLE, Cogview1, Parti}, we generate 30,000 images for the evaluation on MS-COCO~\cite{MSCOCO}. We use metrics\footnote{We use the same evaluation code of prior works~\cite{DALLE, Cogview1} from \href{https://github.com/MinfengZhu/DM-GAN}{https://github.com/MinfengZhu/DM-GAN}} of Fréchet Inception Distance~(FID)~\cite{FID}, Inception Score~(IS)~\cite{IS} and CLIPSIM~\cite{NUWA} to indicate the fidelity, diversity, and text-image relevance, respectively. 

\noindent\textbf{Quantitative Analysis}\quad
\cref{zero-shot} shows the performance comparison of Lformer and other generative transformer T2I models. The Lformer trained on LAION90M achieves an FID score of 15.01 and an IS of 26.22, which is competitive in transformer-based methods. Parti~\cite{Parti} demonstrates that the performance of the generative transformers can be significantly improved as the model scale grows. Thus the tens-billion-scale models are still leading in the FID score. Make-A-Scene~\cite{Make_A_Scene} gets a lower FID score with the help of the segmentation maps thus we do not directly compare with it. 
We also find that the the scale of pre-training dataset plays an important role. The model trained on LAION90M~\cite{BLIP} outperforms the model trained on CC3M by a large margin. Lformer has the potential to get better results if trained with more resources~(both data and computation resources).
Besides, we finetune Lformer on the MS-COCO~\cite{MSCOCO} dataset and report the finetuned FID score. Lformer achieves a finetuned FID score of 9.57, which is better than the finetuned FID score of Cogview2 and NUWA.

\noindent\textbf{Visual Analysis}
Lformer is capable of generating high-quality images with complex text inputs. As shown in \cref{intro_show}, Lformer can generate both photorealistic and artistic content. 
\cref{diversty} shows the diversity of the generated images. 
We also try the combination of a pattern and a carrier~(tote bag and T-shirt) to show the creativity of the model.

\subsection{Image Editing Applications}
\label{sec:edit}
Lformer has the capability to edit images without the requirement of additional fine-tuning. The discrete tokens have strong space-mapping relations with the generated images. Thus we can edit an image by modifying the corresponding tokens of the target area. Based on this, we show two image editing applications: repainting and inpainting.

\begin{figure}[tbp]
    \centering
    \includegraphics[width=0.95\columnwidth]{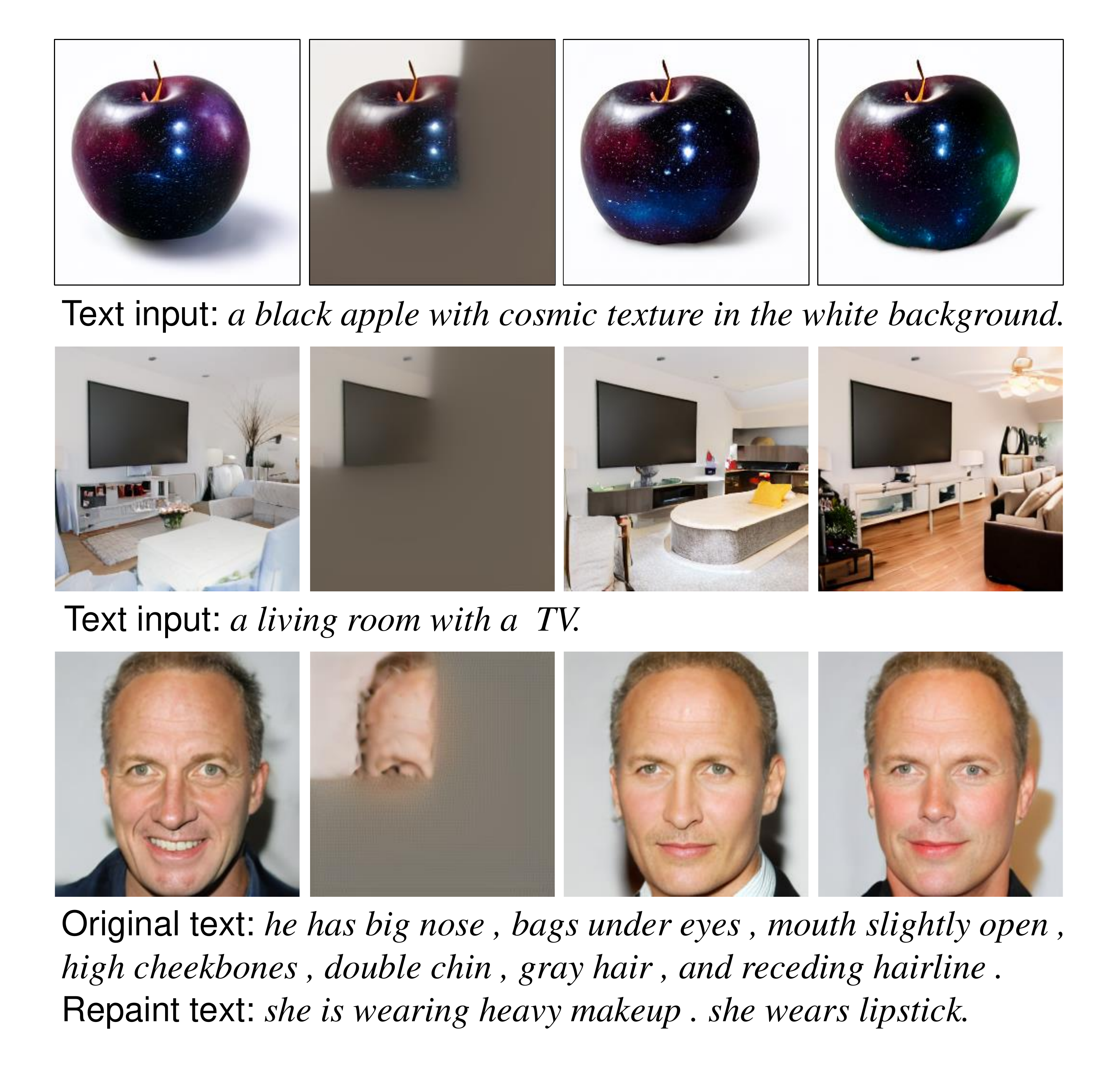}
    \caption{Repainting examples. We can generate variations for an image or achieve the effect of style mixing with different text prompts.}
    \label{repaint}
\end{figure}

\noindent\textbf{Repainting}\quad
Given $h{\times}h$ tokens of an image, we can roll back to the early step $t$ by keeping the early $t{\times}t$ tokens and regenerate the rest of the tokens. \cref{repaint} shows examples of repainting. The first row and the second row show that we can generate variations of an image. In the third row, we generate an image with the description of a male and repaint it with the description of a female. The output image keeps some features of the original image and also introduces the new features of the given text. 

\begin{figure}[tbp]
    \centering
    \includegraphics[width=0.95\columnwidth]{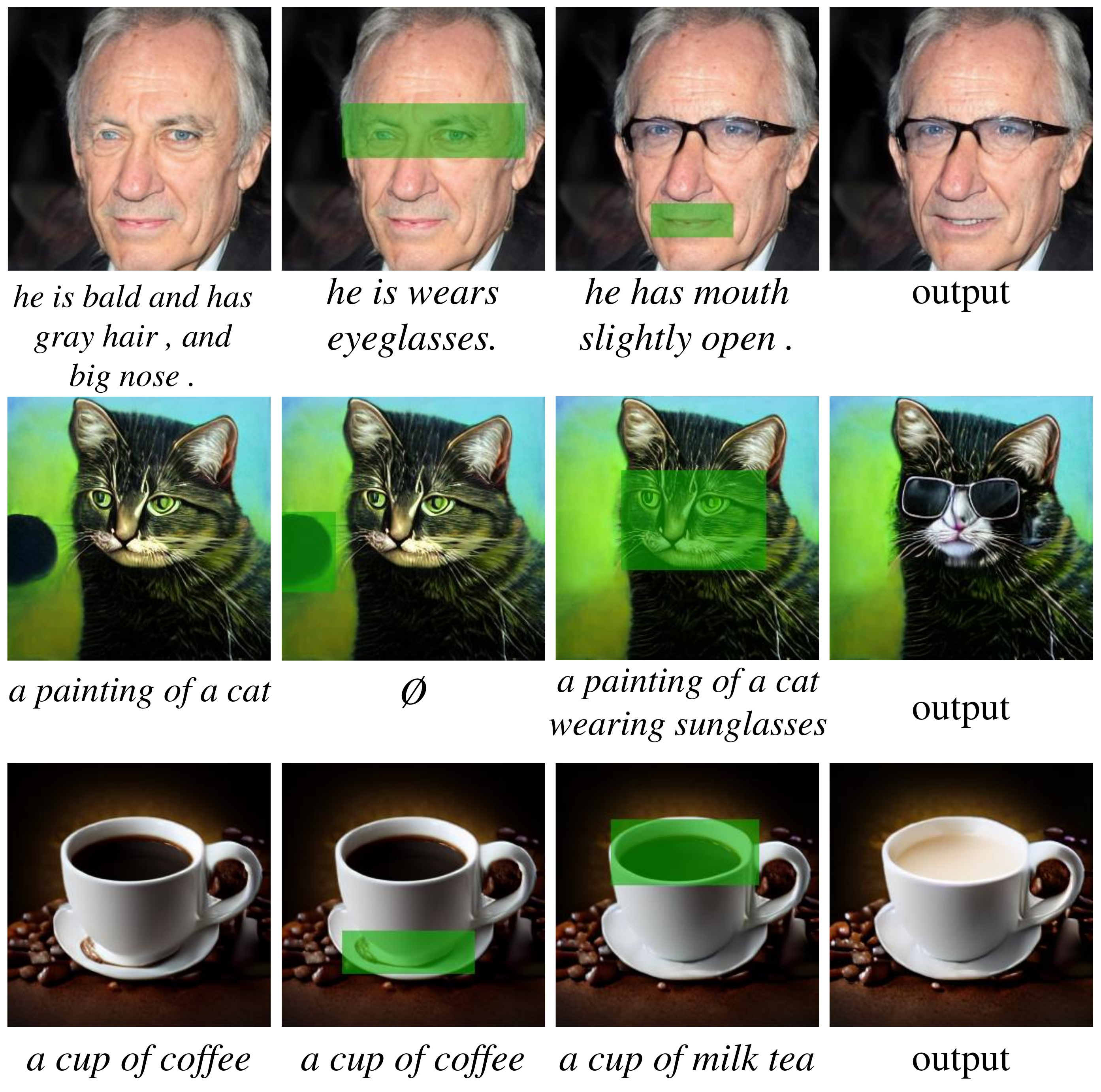}
    \caption{Examples of conditional inpainting with a bounding box and a text prompt. We first generate an image with a text prompt and then repair defects, or edit the image with a different text prompt.}
    \label{inpaint}
\end{figure}

\noindent\textbf{Inpainting}\quad Given a bounding box $(\hat{x_1}, \hat{y_1}, \hat{x_2}, \hat{y_2})$~(coordinate of the top, left, bottom, and right), we map it to the corresponding region~($x_1, y_1, x_2, y_2$) in the token space by $x_1 = {\lfloor}\frac{\hat{x_1}}{f}{\rfloor}, y_1 = {\lfloor}\frac{\hat{y_1}}{f}{\rfloor}, x_2 = {\lceil}\frac{\hat{x_2}}{f}{\rceil}, y_2 = {\lceil}\frac{\hat{y_1}}{f}{\rceil}$, $f$ is the downsampling factor. 
Then we regenerate the L-block containing this region, which is the block generated between the step of [max$(x_1, y_1)$, max$(x_2, y_2)$]. We only replace the tokens in the box region and keep the rest tokens; This replacement is performed immediately in each inpainting step, instead of regenerating all L-blocks and then replacing them. Besides, Lformer also supports conditional inpainting with a different text input. \cref{inpaint} shows examples of inpainting. We can repair local defects or iteratively create an image by continuous inpainting.

\subsection{Conclusion}
In this paper, we propose Lformer, a semi-autoregressive text-to-image generation model that decodes image tokens in an L-shape block parallelly. Lformer achieves a faster speed than the existing transformer-based models while keeping a competitive performance. Lformer can also produce image variations and edit an image with a text prompt.
Lformer currently generates images in a square window. Thus we consider leveraging the sliding window mechanism to generate non-square images in future work. We also consider higher resolution generation by decoding with more tokens or working with an extra upsampling model.

{\small
\bibliographystyle{ieee_fullname}
\bibliography{egbib}
}

\end{document}